\documentclass[myepj-spec,pdftex, iicol]{mySvjour}
\usepackage{graphicx}
\usepackage[pdfpagemode=UseNone]{hyperref}
\usepackage{color}
\usepackage{xspace}
\usepackage[square,sort&compress,numbers]{natbib}   
\usepackage{verbatim}
\usepackage{amsmath,amssymb,amsfonts} % Typical maths resource packages
\usepackage{tabularx}
\usepackage{multicol}
\usepackage{footnote}
\usepackage{listings}
\usepackage[gen]{eurosym}
\usepackage[switch, modulo]{lineno}
\usepackage{longtable}
\usepackage[utf8]{inputenc} % allow utf-8 input
\usepackage[T1]{fontenc}    % use 8-bit T1 fonts
\usepackage{hyperref}       % hyperlinks
\usepackage{url}            % simple URL typesetting
\usepackage{booktabs}       % professional-quality tables
\usepackage{amsfonts}       % blackboard math symbols
\usepackage{nicefrac}       % compact symbols for 1/2, etc.
\usepackage{microtype}      % microtypography
\usepackage{xcolor}         % colors
\usepackage{graphicx}
\usepackage{float}
\usepackage{caption}
\usepackage{subcaption}
\usepackage{amsmath}

\usepackage{lmodern,bm}                
\usepackage[T1]{sansmath} 
\SetMathAlphabet{\mathsfbf}{sans}{\sansmathencoding}{\sfdefault}{bx}{sl}
\usepackage{etoolbox}
\AtBeginEnvironment{sansmath}{}{}{}

\usepackage{lipsum}

\definecolor{darkblue1}{rgb}{0,0,.2}
\definecolor{darkblue}{rgb}{0,0,.2}
\definecolor{darkred}{rgb}{0.5,0,0}
\pagecolor{white} % Background color
\color{black}     % Text color
\hypersetup{breaklinks=true, 
	colorlinks=true, 
	linkcolor=darkblue1, 
	menucolor=darkblue1, 
	urlcolor=darkblue1,
	citecolor=darkblue1,
	pdftitle={},
	pdfauthor={},
	pdfsubject={},
	pdfkeywords={},
	pdfproducer={}
}
\usepackage{siunitx}

\bibstyle{plain}
\begin{document}

			\begin{flushright}
				\normalsize
				%      \today
			\end{flushright}
			
			\vspace{-2cm}
			
			\title{\Large\boldmath Transfer Learning Across Fast- and Full-Simulation Domains in High-Energy Physics
}
\author{Matthias L. Schott$^1$\footnote{corresponding author: mschott@uni-bonn.de}, Lucie Flek$^2$}
\institute{$^1$ Institute of Physics, University of Bonn, Germany, $^2$ Bonn-Aachen International Center for Information Technology (b-it), University of Bonn, Germany}
			
\abstract{Machine-learning models in high-energy physics are often trained on simulated data, where fully simulated samples are computationally expensive while fast simulation provides large statistics at reduced realism. In this work, we systematically study transfer learning between fast-simulated and fully simulated datasets in a realistic LHC environment. 
We consider three representative tasks, signal-background classification, quark-gluon jet tagging, and missing transverse energy reconstruction, using dense neural networks, graph neural networks, and transformer-based architectures. Models are pretrained on ATLAS-like fast simulation and adapted to CMS-like fast simulation and to fully simulated ATLAS Open Data.
Across all tasks, pretrained models consistently outperform independently trained baselines and require significantly less target-domain training data, typically reducing the needed statistics by about a factor of two. These results demonstrate that fast simulation can be used to learn robust, reusable representations and motivate publishing trained models as reusable scientific assets beyond large foundation models.}

\maketitle

\tableofcontents

\section{Introduction and Related Work}

Machine learning (ML) methods have become an integral part of modern high-energy physics (HEP) analyses. From event classification and object identification to regression tasks such as energy and momentum reconstruction, neural networks now routinely outperform traditional, hand-crafted approaches. The works cited in~\cite{Albertsson:2018maf, Radovic:2018dip, Guest:2018yhq, Plehn:2022ftl, Feickert:2021ajf} provide representative entry points into the rapidly growing literature on machine learning in HEP and contain extensive further references covering the breadth of developments in this field. The trend is expected to accelerate as the Large Hadron Collider (LHC) enters higher-luminosity phases, where experimental datasets grow not only in size but also in complexity.

At the same time, the increasing sophistication of detector simulations and physics modeling places significant demands on computing resources, making the efficient use of simulated data a central challenge for current and future analyses. A particularly costly component of the analysis pipeline is the production of fully simulated Monte Carlo samples, typically based on detailed detector descriptions and particle transport using frameworks such as \textsc{Geant4}~\cite{GEANT4:2002zbu}. While these simulations provide the most realistic representation of detector effects, they are computationally expensive and therefore limited in available statistics. In contrast, fast simulation frameworks, such as parametrized detector simulations implemented in \textsc{Delphes}~\cite{deFavereau:2013fsa}, enable the generation of large event samples at a fraction of the cost, but at the price of reduced realism and simplified modeling of detector response and pile-up conditions.

Several research directions have emerged to address the resulting tension between simulation fidelity and available statistics. One prominent approach focuses on improving the realism of fast simulations (\textit{FastSim}) so that they better reproduce the characteristics of full \textsc{Geant4}-based simulations (\textit{FullSim}). For example, \cite{Simsek:2024zhj} introduced \textit{CALPAGAN}, which uses conditional Generative Adversarial Networks (GANs) to translate fast-simulated calorimeter images into higher-fidelity representations. Similarly, the CMS Collaboration has explored regression-based neural networks to provide post-hoc corrections to FastSim observables, improving their agreement with FullSim benchmarks~\cite{Bein:2023ylt}.

A second line of research addresses the gap between simulation and experimental data, commonly referred to as \textit{sim-to-real} transfer. Early studies using CMS Open Data demonstrated that while simulated samples reproduce many jet substructure observables well, measurable discrepancies remain~\cite{Tripathee:2017ybi}. To mitigate such effects, domain adaptation techniques have been explored. For example, \cite{Baalouch:2019fhm} applied Domain Adversarial Neural Networks (DANNs) to learn domain-invariant representations that maintain performance when models trained on simulation are applied to real experimental data.

More recently, the field has begun exploring large-scale pretraining strategies inspired by foundation models. The introduction of the \textit{JetClass} dataset and the \textit{Particle Transformer (ParT)} architecture demonstrated that training on extremely large simulated datasets can significantly improve downstream jet-tagging performance~\cite{Qu:2022mxj}. This paradigm has been further extended by the \textit{OmniLearned} framework~\cite{Bhimji:2025isp}, which pretrains models on more than one billion jets and achieves state-of-the-art performance across multiple tasks and experiments.

In parallel to these developments, transfer learning has emerged as a promising strategy for making more efficient use of available simulation samples. In a broad sense, transfer learning refers to the reuse of models or learned representations from one task or dataset to accelerate learning or improve performance on another related task. In HEP, related ideas have appeared in studies exploring the reuse of jet taggers across phase-space regions, physics processes, or detector configurations~\cite{Shimmin:2017mfk,Kasieczka:2019dbj}. However, systematic investigations of transfer learning across fundamentally different simulation setups, detector descriptions, and physics modeling choices remain limited.

In this work we investigate transfer learning in a controlled yet non-trivial HEP setting, extending earlier studies~\cite{Mokhtar:2025zqs, Kishimoto:2022ibr}. We focus on three representative analysis tasks: binary event classification, quark--gluon jet tagging, and missing transverse energy reconstruction. These tasks span a broad range of machine learning applications in collider physics and utilize well-established architectures, including dense neural networks, graph neural networks for jet tagging~\cite{Battaglia:2018qql,Qu:2019gqs}, and transformer-based approaches for event-level regression~\cite{Mikuni:2021pou,Qu:2022mxj}. Our study spans multiple simulation domains, including fast-simulated samples emulating detector conditions of both the ATLAS~\cite{ATLAS:2008xda} and CMS~\cite{CMS:2008xjf} experiments, as well as fully simulated samples derived from ATLAS Open Data~\cite{ATLAS:OpenData1, ATLAS:OpenData2, Albornoz:2024jbm,Saala:2025gvd}.

Importantly, the differences between these datasets go far beyond detector resolution effects. They also include substantial variations in physics modeling, such as the perturbative order of matrix-element calculations, parton shower configurations, and underlying-event tuning, typically implemented using event generators such as \textsc{Pythia8}~\cite{Sjostrand:2007gs}. This makes the transfer between datasets a genuine stress test of the flexibility and generalization capability of modern neural networks. By studying how models trained in one simulation domain perform when transferred to another, we aim to quantify how effectively transfer learning can reduce the dependence on expensive high-fidelity simulations and improve the reuse of trained models across different experimental environments.

\section{Data Sets}

This study is based exclusively on simulated Monte Carlo event samples designed to probe transfer learning across different detector descriptions, pile-up conditions, and physics modeling choices. Four physics processes are considered in order to cover a representative set of final states and analysis challenges commonly encountered at the LHC, namely top-quark pair production in the semi-leptonic decay channel, single vector boson production in association with jets ($W$+jets and $Z$+jets) in the leptonic decay channel, and diboson $WW$ production in the semi-leptonic decay channel. All samples are generated at a proton--proton center-of-mass energy of $\sqrt{s} = \mathrm{13}$~TeV and correspond to inclusive production within fiducial phase-space selections appropriate for the respective decay channels. Unless stated otherwise, generator-level event weights are not used and all events enter the training with equal weight.

To explicitly study domain shifts and their impact on machine-learning models, three distinct simulation domains are employed. Fast-simulated samples with an ATLAS-like detector configuration are produced using the \texttt{Delphes} framework, in which the detector response is parametrized in terms of object resolutions and efficiencies, and no pile-up. Event generation for these samples is based on \texttt{Pythia8} including its standard underlying-event tuning set, and they serve as the primary source of pretraining for all models studied in this work. A second set of fast-simulated samples is generated using a CMS-like \texttt{Delphes} detector card. Compared to the ATLAS-like fast simulation, both detector resolutions and pile-up settings differ, with an average pile-up of $\langle\mu\rangle = \mathrm{5}$, while the underlying physics generation is again performed using \texttt{Pythia8}. These samples are used to evaluate transfer learning between different detector concepts under otherwise comparable simulation complexity.

In addition to fast simulation, fully simulated samples based on ATLAS Open Data are employed. These samples are produced using a detailed detector simulation based on \texttt{Geant4} and include a realistic modeling of detector material, magnetic fields, and particle interactions. The physics modeling of these samples differs substantially from that of the fast-simulated datasets, including differences in matrix-element calculations, parton shower modeling, and underlying-event tuning. As a result, the transition from Delphes-based fast simulation to ATLAS Open Data represents a significantly larger domain shift than a simple change in detector resolution.

All samples are reconstructed at the analysis level using standard physics objects. Charged-particle tracks are required to satisfy a transverse momentum threshold of $p_{\mathrm{T}} > \mathrm{0.5}$~GeV and are used as low-level inputs for graph-based and transformer-based models. Jets are reconstructed using the anti-$k_{\mathrm{T}}$ algorithm with a radius parameter of $R = 0.4$ and are required to have transverse momentum $p_{\mathrm{T}} > 30$~GeV and pseudorapidity $|\eta| < \mathrm{2.5}$. Isolated charged leptons, either electrons or muons, are used to select leptonic and semi-leptonic final states, depending on the physics process under consideration.

For models operating on variable-length particle collections, such as the graph neural network and the transformer-based architecture, a fixed maximum number of tracks is used as input. Events or jets with fewer tracks are zero-padded accordingly, ensuring a consistent input dimensionality across the dataset.
For each physics process and simulation domain, more than $1 \times 10^{5}$ events are available, ensuring that statistical uncertainties do not dominate the training or evaluation of the models. In the transfer-learning studies, subsets of these samples with varying sizes are used to systematically quantify the dependence of model performance on the amount of available target-domain training data. A summary of all datasets used in this analysis is provided in Table~\ref{tab:datasets}.

\begin{table}[htbp]
\centering
\caption{Summary of the simulated datasets used in this study. Event counts are approximate and may vary slightly between tasks due to selection requirements.}
\label{tab:datasets}
\begin{tabular}{l l l c}
\hline
\textbf{Process} & \textbf{Simulation Type} & \textbf{Detector Model} & \textbf{Events} \\
\hline
$t\bar{t}$ (semi-leptonic) & Fast simulation & \textsc{Pythia8}, Delphes (ATLAS-Card) & $100\,000$ \\
$t\bar{t}$ (semi-leptonic) & Fast simulation & \textsc{Pythia8}, PileUp, Delphes (CMS-Card)   & $100\,000$ \\
$t\bar{t}$ (semi-leptonic) & Full simulation & \textsc{PowhegPythia}, PileUp, Geant4 (ATLAS Open Data \cite{ATLAS:OpenData2}) & $150\,000$ \\
\hline
$W$+jets (leptonic)     & Fast simulation & \textsc{Pythia8}, Delphes (ATLAS-Card) & $200\,000$ \\
$W$+jets (leptonic)     & Fast simulation & \textsc{Pythia8}+PileUp Delphes (CMS-Card) & $200\,000$ \\
$W$+jets (leptonic)     & Full simulation & \textsc{Sherpa}, PileUp, Geant4 (ATLAS Open Data \cite{ATLAS:OpenData1}) & $150\,000$ \\
\hline
\hline
$Z$+jets (leptonic)     & Fast simulation & \textsc{Pythia8}, Delphes (ATLAS-Card) & $100\,000$ \\
$Z$+jets (leptonic)     & Fast simulation & \textsc{Pythia8}+PileUp Delphes (CMS-Card) & $100\,000$ \\
$Z$+jets (leptonic)     & Full simulation & \textsc{Sherpa}, PileUp, Geant4 (ATLAS Open Data \cite{ATLAS:OpenData1}) & $150\,000$ \\
\hline
$WW$ (semi-leptonic)      & Fast simulation & \textsc{Pythia8}, Delphes (ATLAS-Card) & $100\,000$ \\
$WW$ (semi-leptonic)      & Fast simulation & \textsc{Pythia8}, PileUp, Delphes (CMS-Card)   & $100\,000$ \\
$WW$ (semi-leptonic)      & Full simulation & \textsc{Sherpa}, PileUp, Geant4 (ATLAS Open Data \cite{ATLAS:OpenData1})& $200\,000$ \\
\hline
\end{tabular}
\end{table}

The ATLAS Delphes samples are used as the default source for model pretraining, while CMS Delphes and ATLAS Open Data samples serve as target domains for transfer-learning studies with varying amounts of retraining data.

\section{Hypothesis and Models}

The central hypothesis of this work is that neural networks pretrained on one simulation domain can be successfully transferred to a different domain while significantly reducing the required size of the target training dataset, yet achieving performance comparable to—or better than—models trained from scratch. In particular, we hypothesize that the advantages of transfer learning are already present in relatively small-scale architectures and are not limited to highly complex, large transformer-based models. In the context of high-energy physics, this hypothesis directly addresses the practical challenge that realistic, fully simulated datasets are computationally expensive and therefore limited in size, whereas fast-simulated datasets are abundant but less accurate.

More specifically, this study tests the assumption that pretrained models require substantially fewer target-domain events to reach a given performance level than models trained independently from scratch. It further examines whether allowing all network parameters to be retrained provides sufficient flexibility to adapt to large domain shifts, including changes in detector simulation and physics modeling. Finally, the potential benefit of freezing early network layers is investigated, with the expectation that this strategy can be advantageous when the domain shift is moderate, but may reduce performance when the differences between source and target domains become large.

If these hypotheses hold, several important implications follow. First, fast simulation could be used more systematically to learn robust and reusable representations, which can later be adapted to fully simulated samples with limited additional cost. Second, results obtained in feasibility studies or exploratory analyses based on fast simulation would gain additional credibility, as pretrained models could be validated and refined using small amounts of high-fidelity simulation. Finally, this would motivate the publication of trained models and pretrained weights not only for large foundation models, but also for task-specific architectures commonly used in HEP analyses.

To test these hypotheses in a comprehensive manner, we consider three representative machine-learning tasks that cover the most common use cases in current LHC analyses: event-level classification, jet-level tagging, and event-level regression. These tasks are deliberately chosen to span a wide range of input representations and neural network architectures, including dense neural networks (DNNs), graph neural networks (GNNs), and transformer-based models.

The analysis presented in this work is based on three complementary machine-learning tasks that together cover the most common use cases in contemporary high-energy physics analyses. The first task is \textbf{signal--background classification}, formulated as a binary classification problem in which top-quark pair events are distinguished from diboson $WW$ events in the semi-leptonic decay channel using high-level kinematic observables. The second task is \textbf{quark--gluon jet tagging}, a jet-level classification problem that relies on low-level charged-particle track information and is implemented using a graph neural network to capture the internal structure of jets. The third task is \textbf{missing transverse energy reconstruction}, an event-level regression problem in which a transformer-based architecture is used to estimate the missing transverse energy from track-level inputs. The model structures are detailed in the following. 

\subsection{Signal-Background Classification Model}

Signal–background classification is one of the most fundamental and widely used tasks in LHC data analyses. It forms the backbone of a broad range of measurements and searches, where the primary objective is to separate a physics process of interest from often overwhelming backgrounds. Typical examples include the discrimination of Higgs boson production from Standard Model background processes (e.g. \cite{ATLAS:2024jry, CMS:2025ihj}), the separation of top-quark pair production from electroweak backgrounds (e.g. \cite{CMS:2025kzt, ATLAS:2024hac, Andrews:2021ejw}), searches for rare or exotic signals with signatures similar to well-known processes (e.g. \cite{CMS:2024zpb, ATLAS:2023ian}), and precision measurements where background contamination must be tightly controlled. Over the years, this task has been addressed using a wide variety of approaches, ranging from simple cut-based selections to multivariate techniques and, more recently, deep neural networks operating on both high-level and low-level observables.
Given the ubiquity of signal–background classification at the LHC, the choice of a specific example in this study is necessarily representative rather than exhaustive. The goal here is not to cover the full diversity of classification problems encountered in collider physics, but instead to select a well-defined and realistic benchmark that captures the essential characteristics of many analyses. By focusing on a standard signal–background classification task with commonly used input features and a simple yet effective neural network architecture, this study provides a controlled setting in which the impact of transfer learning can be systematically investigated and the results can be readily interpreted and generalized to a wide range of similar applications.

For the representative signal--background classification task, a fully connected deep neural network is employed. The input to the network consists of 12 high-level kinematic features derived from the two leading jets, the reconstructed charged lepton kinematics, and the total number of jets with transverse momentum above 40~GeV. These variables are representative of typical analysis-level observables used in LHC measurements and provide a compact, low-dimensional description of the event.

\begin{figure}[htbp]
    \centering
    \includegraphics[width=0.32\textwidth]{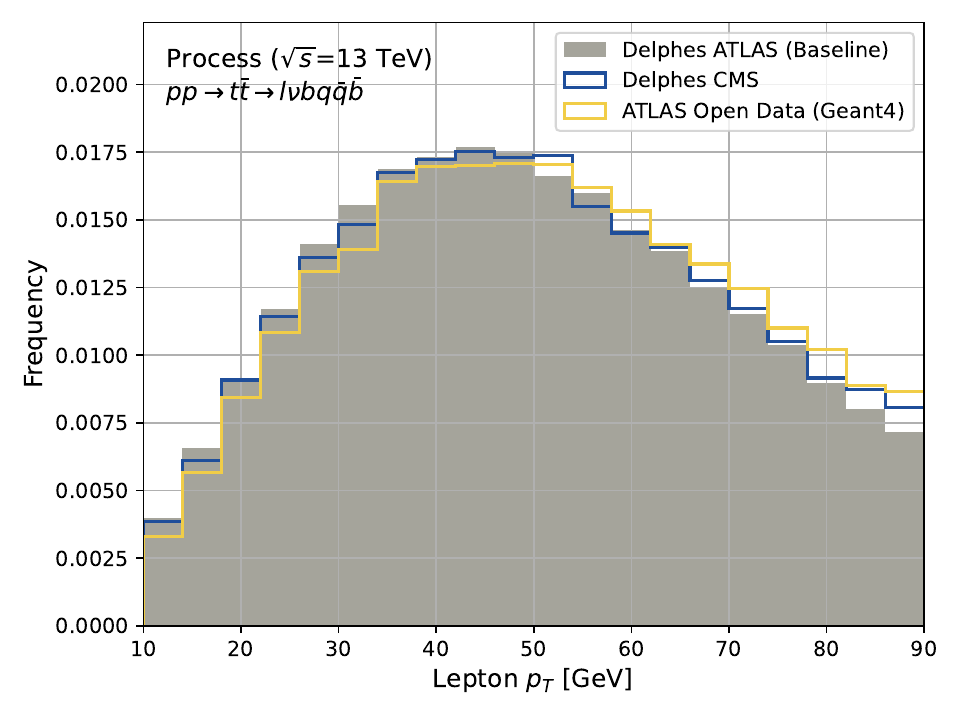}
    \includegraphics[width=0.32\textwidth]{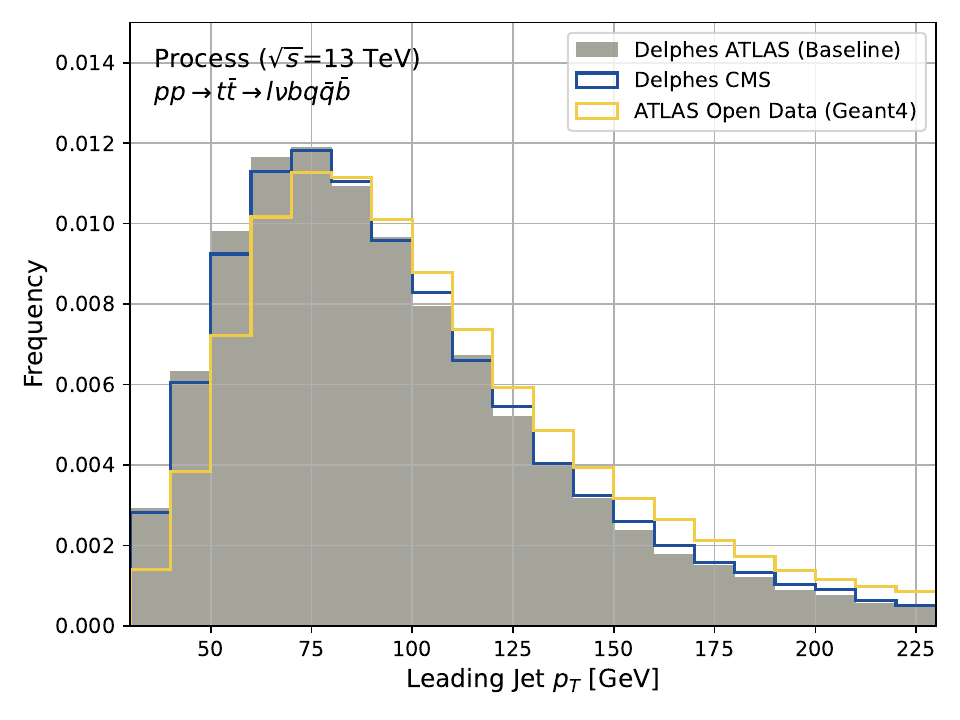}
    \includegraphics[width=0.32\textwidth]{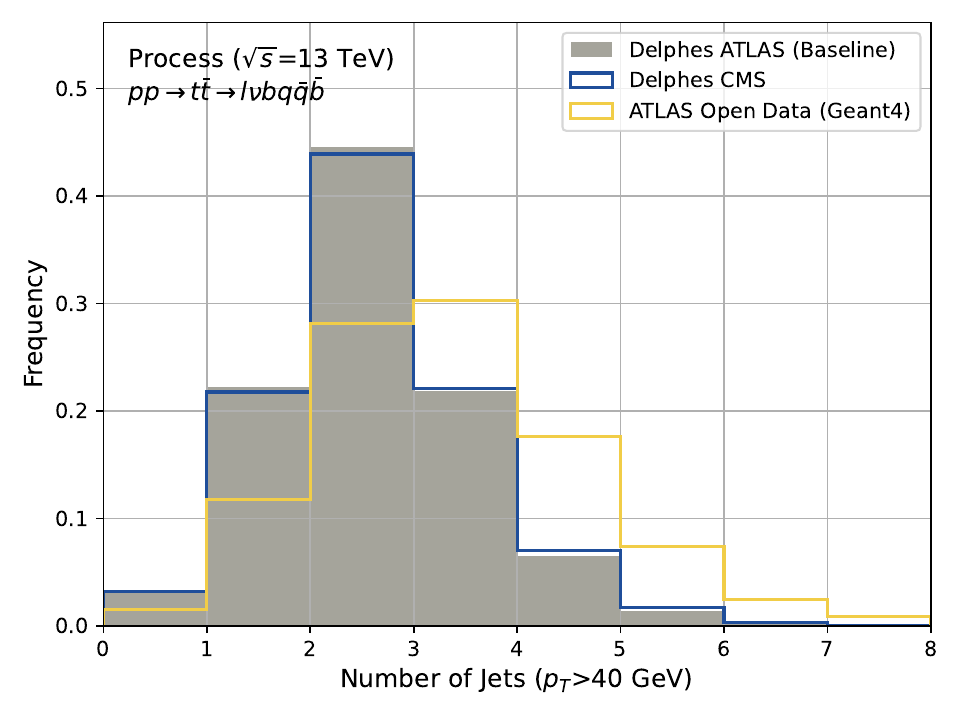}
    \caption{Comparison of three representative input features for the Signal/background classification task, namely the lepton transverse momentum $p_{\mathrm{T}}$ (left), leading-jet transverse momentum $p_{\mathrm{T}}$ (middle), and the number of reconstructed jets (right). The ATLAS Delphes fast-simulated samples used as the baseline are indicated in gray, the CMS Delphes fast-simulated samples in blue, and the fully simulated ATLAS Open Data samples in yellow.}
    \label{fig:featuresWWTop}
\end{figure}

Three representative examples of input feature distributions for the top-quark pair process, utilized for the Signal–background classification task, are shown in Figure~\ref{fig:featuresWWTop}, namely the transverse momentum of the reconstructed lepton, the transverse momentum of the leading jet, and the number of reconstructed jets. The distributions are compared for fast-simulated samples using ATLAS- and CMS-like detector configurations as well as for the fully simulated ATLAS Open Data sample. As expected, only relatively small differences are observed between the two fast-simulation setups, since these variations originate primarily from differences in detector resolutions and reconstruction efficiencies. In contrast, significantly larger discrepancies are observed when comparing the ATLAS fast-simulation samples to the fully simulated ATLAS Open Data samples. These differences reflect not only the more detailed detector description provided by the Geant4-based simulation, but also substantial changes in the underlying physics modeling. In particular, the fully simulated samples exhibit harder jet transverse momentum spectra and a higher jet multiplicity, which can be attributed to the inclusion of higher-order matrix-element calculations and a more sophisticated treatment of additional QCD radiation.

The network architecture is intentionally kept simple and is composed of a sequence of linear layers with rectified linear unit (ReLU) activations and dropout regularization. Starting from the 12-dimensional input layer, the network comprises two hidden layers with 64 neurons each, followed by a hidden layer with 32 neurons, and a single-node output layer with a sigmoid activation that produces a binary classification probability. Dropout is applied after each hidden layer to mitigate overfitting. Despite its simplicity, this architecture achieves strong discrimination performance and serves as a baseline example for studying transfer learning in the context of tabular input data.

\subsection{Quark-Gluon Jet Tagging Model}

Quark–gluon jet tagging is a long-standing and scientifically important problem in collider physics, with direct relevance for both precision measurements and searches for new phenomena at the LHC. The ability to distinguish jets initiated by quarks from those initiated by gluons can improve background rejection in many analyses and even more importantly in SM precision measurements. Early approaches relied on hand-crafted observables based on jet shapes and particle multiplicities \cite{Cornelis:2014ima}, while more recent studies have demonstrated substantial performance gains using machine-learning techniques that operate directly on low-level jet constituents, including deep neural networks and set-based architectures \cite{Andrews:2019faz}. Notable examples include energy-flow–based methods and particle-cloud representations, which have established low-level learning as a powerful paradigm for jet tagging \cite{Qu:2019gqs}. Graph neural networks are a particularly natural choice for quark–gluon discrimination, as jets can be interpreted as unordered collections of particles with non-trivial geometric and relational structure in momentum space. By explicitly modeling local correlations between nearby particles and aggregating this information into global jet representations, GNNs provide a flexible and physically motivated framework that has been shown to achieve state-of-the-art performance while remaining robust to variations in detector resolution and particle multiplicity.

The quark-gluon jet tagging task is therefore formulated in this work as a graph-learning problem, reflecting the variable number and spatial structure of particle tracks inside a jet. Each jet is represented by up to 50 charged-particle tracks, described by their transverse momentum, pseudorapidity, azimuthal angle, charge, transverse impact parameter $d_{0}$, and longitudinal impact parameter $z_{0}$. Jets with fewer than 50 tracks are zero-padded.

A graph is constructed dynamically for each jet using a $k$-nearest-neighbor algorithm in the $(\eta,\phi)$ plane, with proper handling of the periodicity in the azimuthal angle. Nodes correspond to tracks, and edges encode local geometric proximity within the jet.

The neural network architecture is based on a stack of EdgeConv layers, each followed by nonlinear activation, batch normalization, and dropout. Three EdgeConv blocks are used, with increasing feature dimensions, allowing the network to learn progressively more abstract representations of local and global jet substructure. The outputs of all EdgeConv layers are concatenated and aggregated using a global mean pooling operation to form a fixed-size jet representation. This representation is passed to a fully connected classification head consisting of two hidden layers and a final linear output producing a single logit. This architecture captures both local correlations between nearby tracks and global properties of the jet, making it a standard and powerful approach for particle-level jet tagging.

\begin{figure}[htbp]
    \centering
    \includegraphics[width=0.32\textwidth]{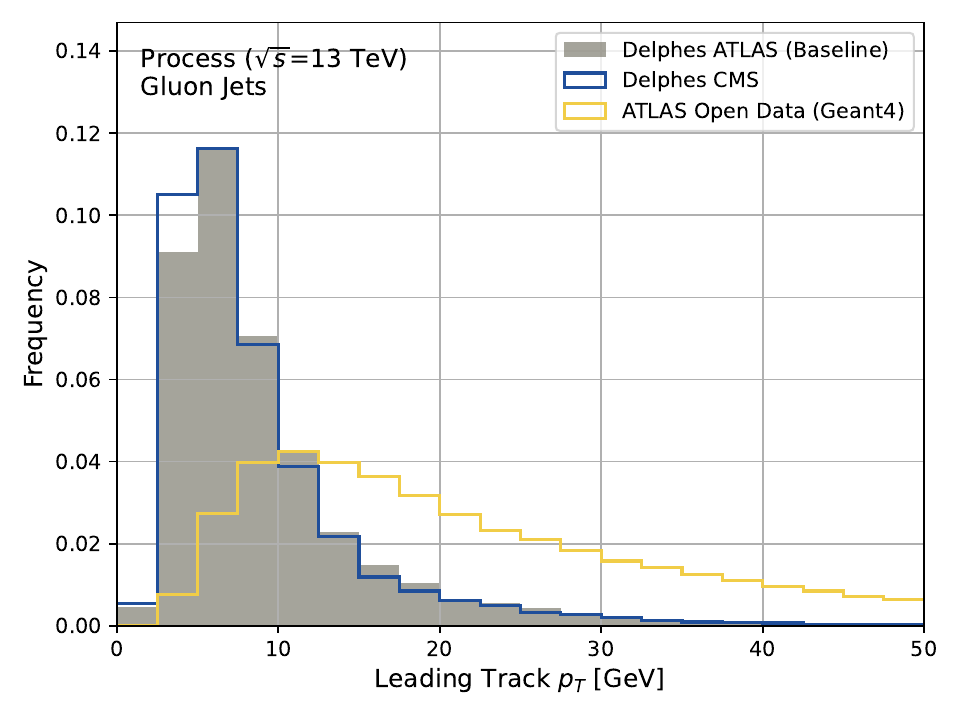}
    \includegraphics[width=0.32\textwidth]{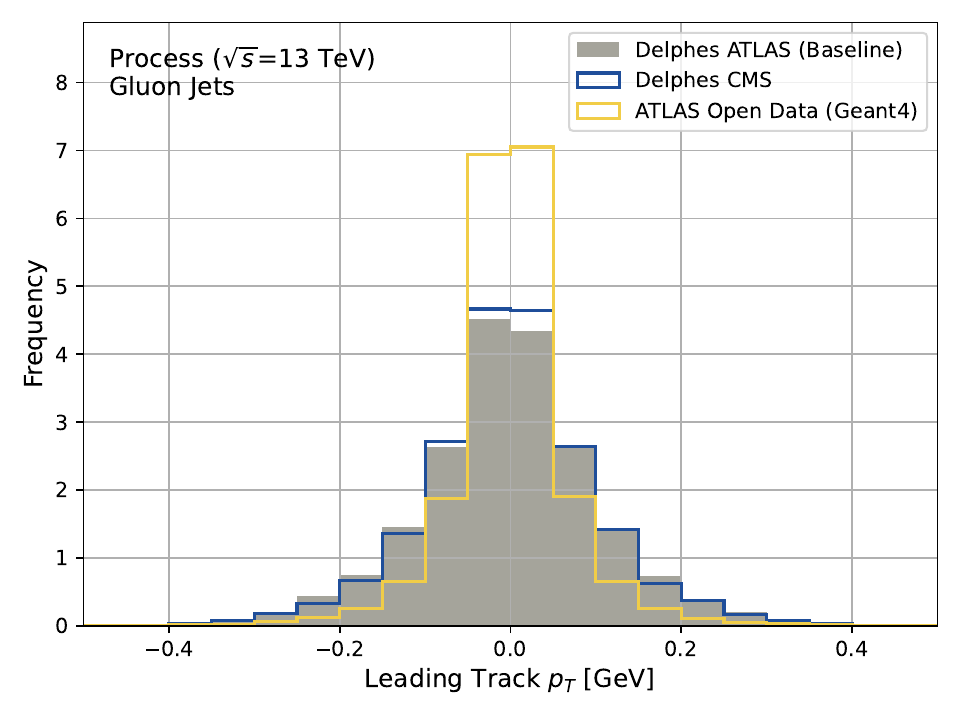}
    \includegraphics[width=0.32\textwidth]{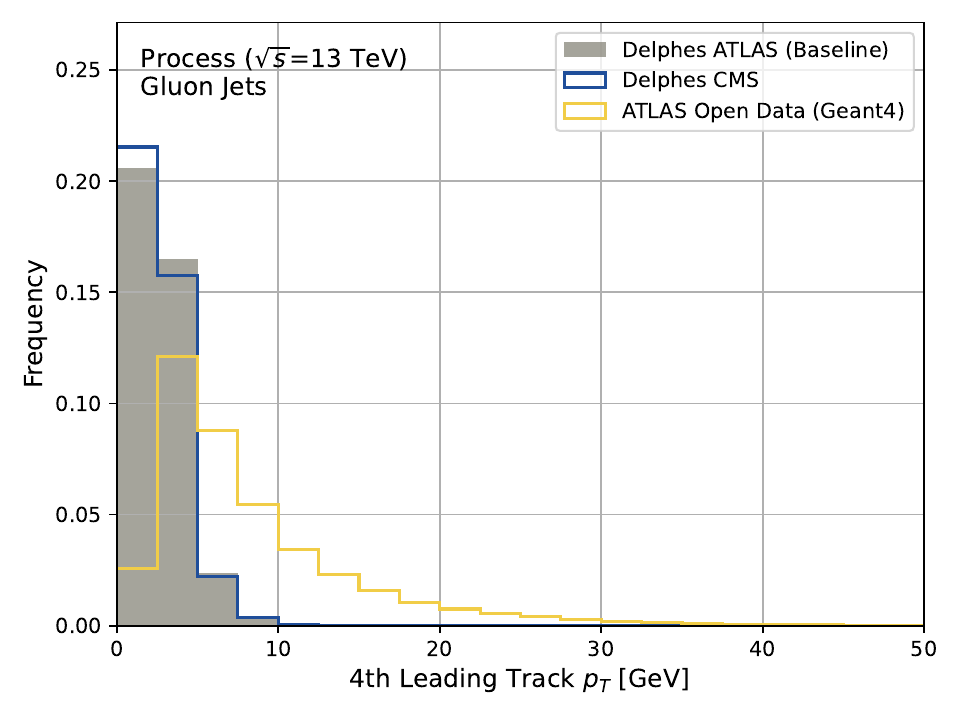}
    \caption{Comparison of three representative input features for the gluon jets, namely the transverse momentum $p_{\mathrm{T}}$ of the leading track (left), the pseudo-rapidity difference to the jet-center of the leading track (middle), and the transverse momentum of the 4th leading track (right). The ATLAS Delphes fast-simulated samples used as the baseline are indicated in gray, the CMS Delphes fast-simulated samples in blue, and the fully simulated ATLAS Open Data samples in yellow.}
    \label{fig:featureJetTagging}
\end{figure}

The training examples for the quark–gluon jet tagging task are based on $W/Z$+jet samples, which are reweighted such that the transverse momentum spectra of quark- and gluon-initiated jets match.  This avoids trivial discrimination based on kinematic differences and ensures that the classification task probes genuine differences in jet substructure. The origin of each jet, namely whether it is initiated by a quark or a gluon, is determined at the Monte Carlo truth level by identifying the highest-momentum parton within a defined vicinity of the reconstructed jet. Three representative input features for gluon jets are shown in Figure~\ref{fig:featureJetTagging}: the transverse momentum $p_{\mathrm{T}}$ of the leading track (left), the pseudorapidity difference between the leading track and the jet axis (middle), and the transverse momentum of the fourth-leading track (right). The distributions are compared for ATLAS Delphes fast-simulated samples, which serve as the baseline and are shown in gray, CMS Delphes fast-simulated samples in blue, and the fully simulated ATLAS Open Data samples in yellow.

As in the signal–background classification study, only relatively small differences are observed between the two fast-simulation setups, reflecting the fact that they differ primarily in detector resolution and reconstruction details. In contrast, significantly larger differences are observed when comparing the fast-simulated samples to the fully simulated ATLAS Open Data samples, which generally exhibit harder spectra. This behavior can again be attributed to differences in the underlying physics modeling, in particular the inclusion of higher-order matrix-element contributions in the simulation of $W$ and $Z$ boson production, which leads to enhanced hard radiation and modified jet substructure.

\subsection{Missing Transverse Energy Reconstruction Model}

The third task addresses the reconstruction of missing transverse energy, a key observable in many LHC analyses and a particularly challenging quantity to model accurately. Its reconstruction is sensitive to detector effects, pile-up, and soft radiation, making it an ideal benchmark for testing the robustness and generalization capabilities of machine-learning models. Traditional approaches rely on vectorial sums of reconstructed objects combined with calibration and pile-up mitigation techniques, while more recent neural-network–based methods have explored the use of low-level detector information to improve resolution and reduce biases. Previous studies have employed among others convolutional and attention-based architectures (e.g. \cite{CMS:2025prt, ATLAS:2024cmj, Maier:2021ymx}) to regress missing transverse energy or its components, demonstrating clear improvements over classical algorithms.
In this work, the missing transverse energy reconstruction task is formulated as a regression problem using low-level charged-particle track information, in particular the three momentum vector components $p_x, p_y$ and $p_z$ as well as the impact parameter $d_0$. A transformer-based architecture is employed to process variable-length sequences of reconstructed tracks, enabling the model to capture global event-level correlations through self-attention. Each track is described by four input features, including momentum components and impact parameter information, providing a compact yet expressive representation of the charged-particle activity in the event. This choice of architecture allows the model to naturally handle variable numbers of tracks and to learn long-range dependencies that are expected to be relevant for accurate missing transverse energy reconstruction.

Each track information ($p_x, p_y, p_z, d_0$) is first embedded into a fixed-dimensional latent space using a linear projection followed by a GELU activation and layer normalization. A learnable classification token is prepended to the sequence, enabling the model to aggregate global event information. The embedded sequence is then processed by a stack of transformer encoder layers with multi-head self-attention and feed-forward sublayers.

The output corresponding to the classification token is passed through a regression head consisting of normalization, fully connected layers, and dropout. The model predicts a single continuous quantity related to the missing transverse energy, and is trained using a loss function that enforces both small resolution and minimal bias. This transformer-based model represents the most flexible and expressive architecture considered in this study and serves as an example for transfer learning in complex, set-based regression tasks.

\begin{figure}[htbp]
    \centering
    \includegraphics[width=0.32\textwidth]{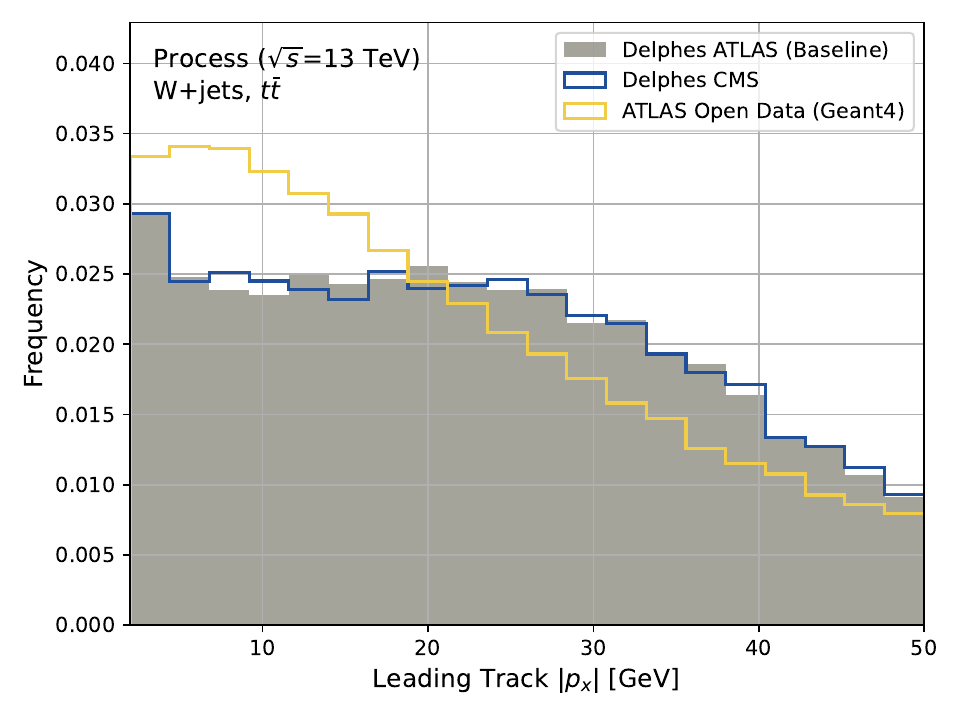}
    \includegraphics[width=0.32\textwidth]{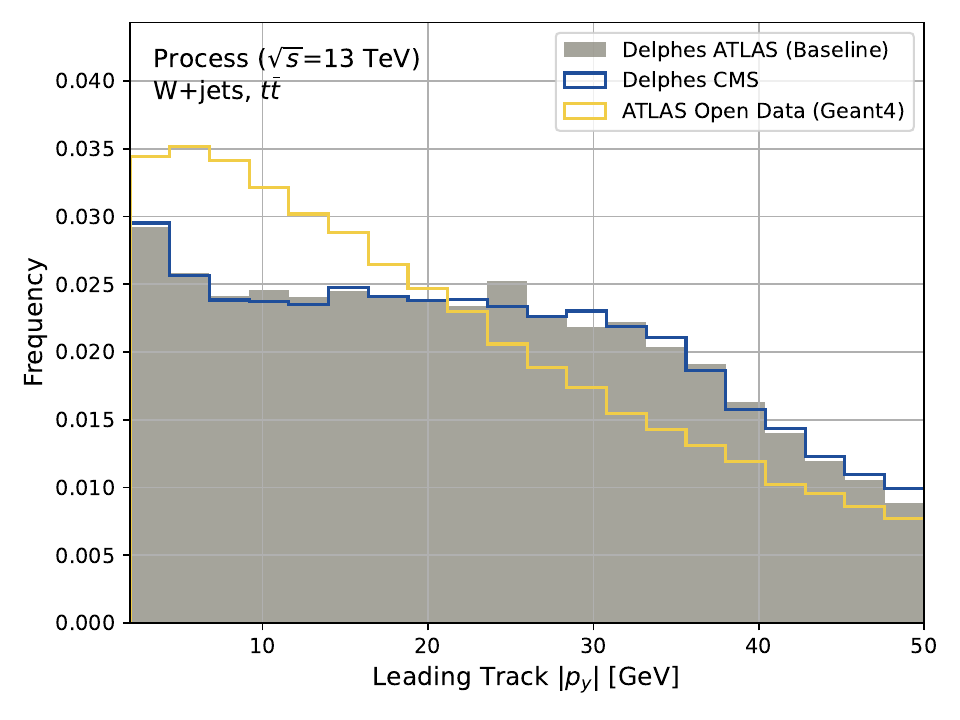}
    \includegraphics[width=0.32\textwidth]{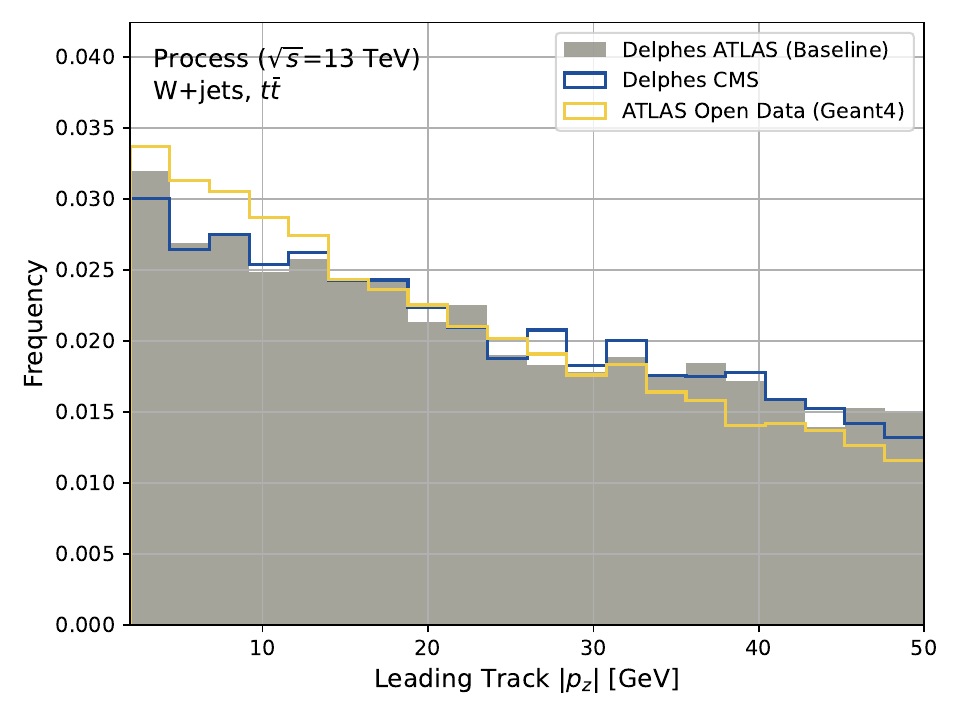}
    \caption{Comparison of three representative input features for the $E_T^{Miss}$ regression task, namely the absolute values of the x- (left), y-( middle) and z-component (right) of the leading track. The ATLAS Delphes fast-simulated samples used as the baseline are indicated in gray, the CMS Delphes fast-simulated samples in blue, and the fully simulated ATLAS Open Data samples in yellow. The shaded bands indicate the statistical uncertainties.}
    \label{fig:featureETMiss}
\end{figure}

As an illustration of the input feature distributions used for the missing transverse energy reconstruction task, the momentum components $|p_x|$, $|p_y|$, and $|p_z|$ of the leading track are shown in Figure~\ref{fig:featureETMiss} for ATLAS Delphes fast-simulated samples, which serve as the baseline, for CMS Delphes fast-simulated samples, and for the fully simulated ATLAS Open Data samples. The distributions are shown for events from the $W$+jets and $t\bar{t}$ processes. As in the previous examples, only modest differences are observed between the two fast-simulation setups, while significantly larger discrepancies appear when comparing the fast-simulated samples to the fully simulated data. These differences again reflect the combined impact of the more detailed detector simulation and the altered underlying physics modeling in the fully simulated samples.

\subsection{Transfer Learning Strategy}

For all three tasks considered in this study, the models are first trained on fast-simulated ATLAS-like samples to obtain a set of pretrained network parameters. For this baseline training, the full available ATLAS fast-simulation dataset is used in each case, ensuring that the pretrained models represent the best possible performance achievable in the source domain and provide a robust starting point for subsequent transfer-learning studies. These pretrained models are then adapted to new simulation domains using two conceptually different strategies that probe the role of model flexibility during transfer.

In the first strategy, all network parameters are left unfrozen and optimized on the target-domain data, allowing the model to fully adapt its internal representations to the new detector description and physics modeling. In the second strategy, only the later layers of the network, or task-specific output heads, are retrained, while the early layers are kept fixed. This partial retraining approach is intended to preserve representations learned in the source domain and to assess whether these representations remain useful when transferring to a different simulation environment.

To quantitatively evaluate the benefit of pretraining, the retrained models are compared to baseline models that are randomly initialized and trained from scratch using only data from the target domain. For each target domain, the retraining and baseline training are performed using progressively increasing subsets of the available training data, allowing the performance to be studied as a function of the target-domain training sample size. This procedure directly probes how much training data is required to reach a given level of performance with and without pretraining.

For each configuration and training sample size, the training is repeated multiple times with different random initializations of the network parameters as well as the training-, validation and test-sets. The spread of the resulting performance metrics is used as an estimate of the statistical uncertainty associated with each experiment. By comparing the performance of pretrained and independently trained models across tasks, architectures, simulation domains, and training statistics, this study assesses how architectural complexity and representational flexibility influence the effectiveness of transfer learning in high-energy physics applications.

\section{Results}

In this section, we present the results of the transfer-learning studies for the three machine-learning tasks introduced in Section~3. For each task, model performance is evaluated as a function of the available training statistics in the target domain and compared to a baseline model trained independently from scratch. A central goal of this section is not only to demonstrate absolute performance improvements, but also to quantify how much target-domain training data can be saved by using pretrained models and how this depends on the choice of retraining strategy.

\subsection{Signal-Background Classification}

The signal-background classification task provides the simplest test case, both in terms of input dimensionality and network architecture. The dense neural network described in Section~3 is trained to distinguish $t\bar{t}$ from $WW$ events in the semi-leptonic channel using high-level kinematic observables.

\begin{figure}[htbp]
    \centering
    \begin{subfigure}[t]{0.48\textwidth}
        \centering
        \includegraphics[width=\textwidth]{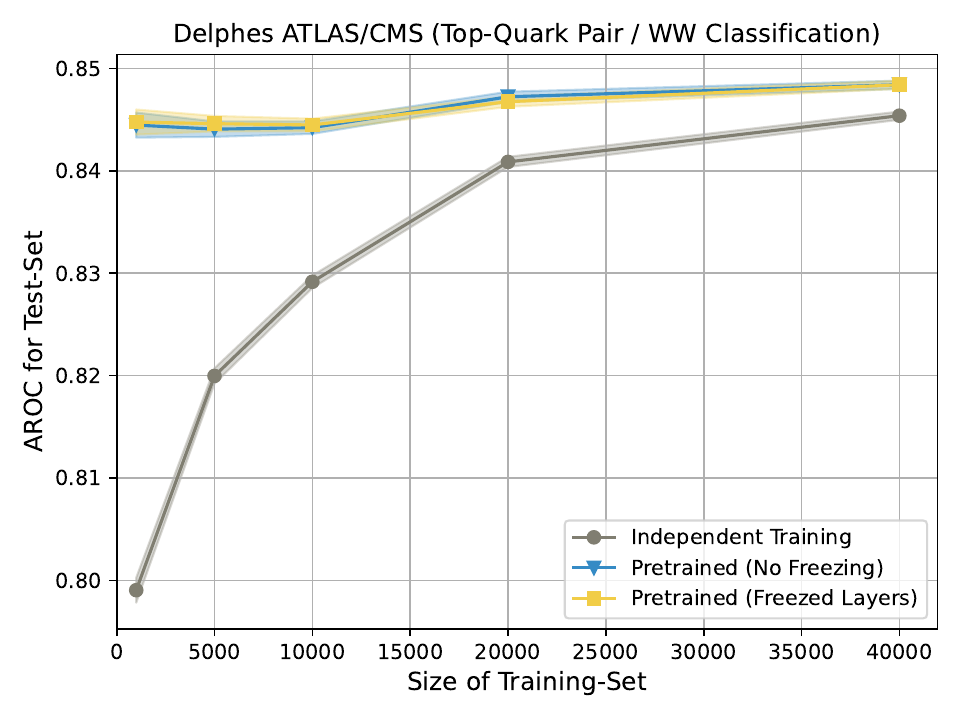}
        \caption{Delphes CMS Simulation.}
        \label{fig:WWTopResults1}
    \end{subfigure}
    \hfill
    \begin{subfigure}[t]{0.48\textwidth}
        \centering
        \includegraphics[width=\textwidth]{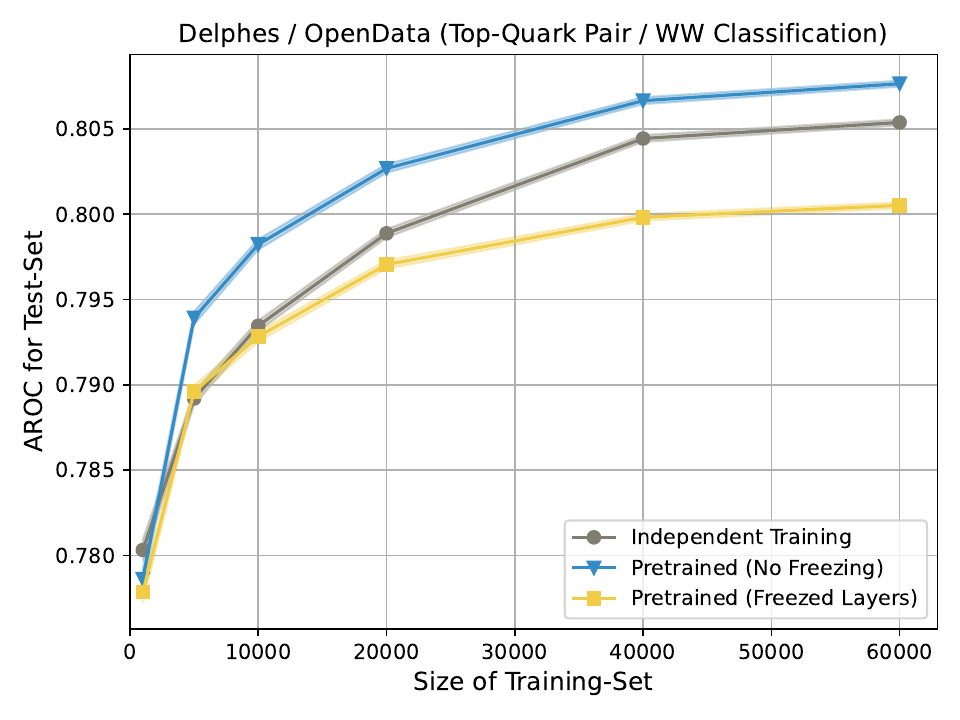}
        \caption{ATLAS Open Data Simulation.}
        \label{fig:WWTopResults2}
    \end{subfigure}
    \caption{Network performance quantified by the area under the ROC curve as a function of the training set size. The gray curve shows the performance of a network trained from scratch with randomly initialized weights. The blue curve corresponds to a network pretrained on the ATLAS Delphes samples, where all network parameters are retrained on the target-domain data. The yellow curve shows the performance of the pretrained network when the early layers are frozen and only the later layers are updated during retraining. The shaded bands indicate the statistical uncertainties, which are generally small compared to the observed effects.}
    \label{fig:WWTopResults}
\end{figure}

When transferring the pretrained model to CMS Delphes samples, a clear and consistent improvement over the independently trained baseline is observed across the full range of training statistics. As illustrated by the ROC curves summarized in Figure~\ref{fig:WWTopResults1}, both transfer-learning strategies outperform the baseline model. The strongest gains are achieved when all network parameters are retrained, indicating that even for relatively similar fast-simulation domains, allowing full flexibility is beneficial. Notably, with only a few thousand CMS Delphes events, the pretrained model already exceeds the performance of a baseline model trained on several tens of thousands of events. This demonstrates that the pretrained network has learned a representation that is highly reusable across detector configurations and pile-up conditions.

Freezing the first layer leads to a modest reduction in performance compared to full retraining, but still provides a significant advantage at low statistics. This suggests that some of the low-level feature representations learned from ATLAS Delphes generalize well to CMS Delphes, while deeper layers benefit from adaptation.

The situation changes when transferring to ATLAS Open Data full simulation, as shown in Figure~\ref{fig:WWTopResults2}. Here, the independently trained baseline improves steadily with increasing training statistics, but the pretrained model with frozen layers consistently underperforms, indicating that the constraints imposed by freezing the early layers are too restrictive given the large domain shift. In contrast, the pretrained model with full retraining remains superior to the baseline for all training set sizes considered. Comparable performance is achieved with roughly half of the training data, despite the substantially different detector simulation and physics modeling. 

\subsection{Quark-Gluon Jet Tagging}

The quark-gluon jet tagging task represents a more complex and realistic use case, relying on low-level track information and a graph neural network architecture. In this setting, transfer learning must cope not only with detector effects but also with changes in track multiplicities, resolutions, and jet substructure.

For the transfer from ATLAS Delphes to CMS Delphes, the performance trends shown in Figure~3 closely mirror those observed in the simpler classification task. The pretrained model with all layers retrained consistently outperforms the independently trained baseline across all training statistics. Again, approximately half of the training data is sufficient to reach a similar ROC performance compared to training from scratch.

The behavior of the partially frozen model is particularly instructive in this case. At very low statistics, freezing the GNN layers yields a noticeable advantage over both the baseline and the fully retrained model, indicating that the learned local jet substructure representations are robust against moderate domain shifts. However, as the training set size increases, the performance saturates and converges to that of the independently trained model. This suggests that while frozen representations are helpful as a strong inductive bias in the low-statistics regime, they ultimately limit the achievable performance once sufficient data is available.

\begin{figure}[htbp]
    \centering
    \begin{subfigure}[t]{0.48\textwidth}
        \centering
        \includegraphics[width=\textwidth]{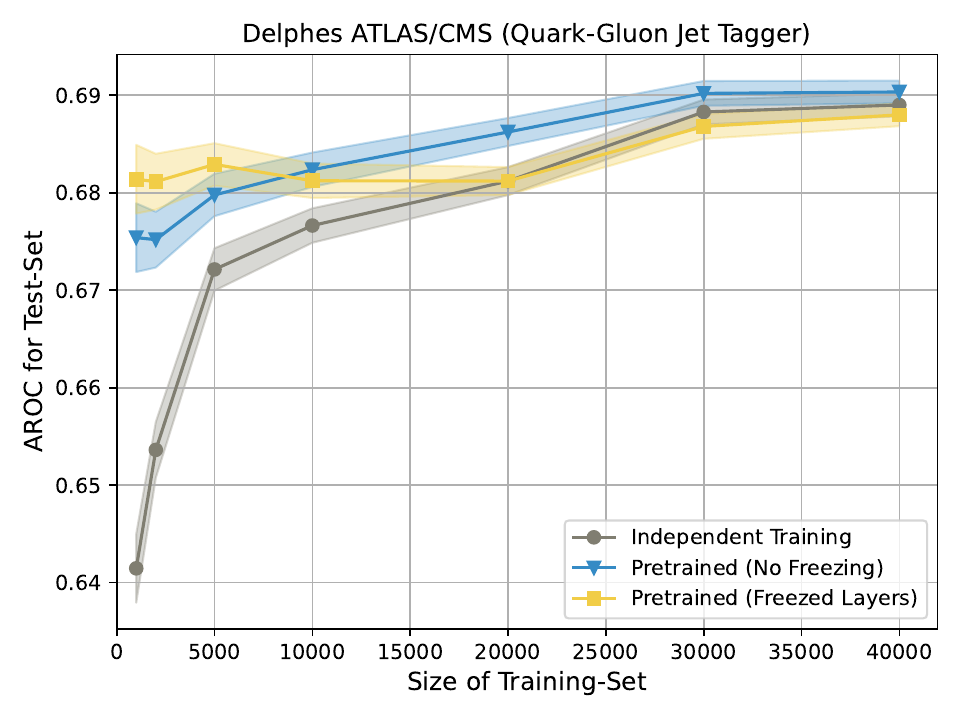}
        \caption{Delphes CMS Simulation.}
        \label{fig:JetTaggerResults1}
    \end{subfigure}
    \hfill
    \begin{subfigure}[t]{0.48\textwidth}
        \centering
        \includegraphics[width=\textwidth]{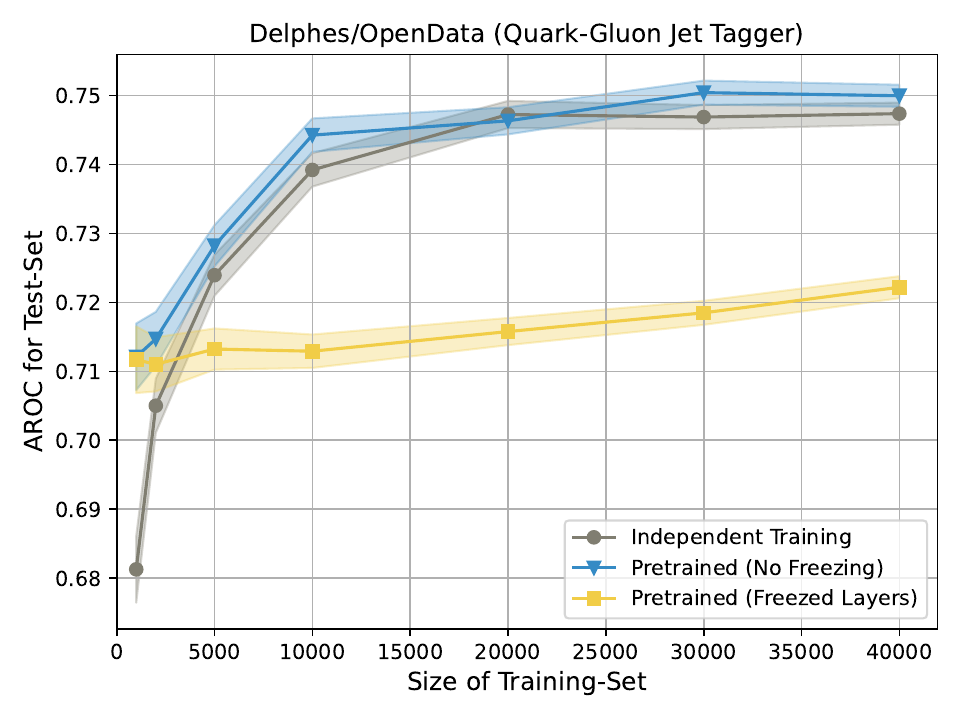}
        \caption{ATLAS Open Data Simulation.}
        \label{fig:JetTaggerResults2}
    \end{subfigure}
    \caption{Network performance quantified by the area under the ROC curve as a function of the training set size. The gray curve shows the performance of a network trained from scratch with randomly initialized weights. The blue curve corresponds to a network pretrained on the ATLAS Delphes samples, where all network parameters are retrained on the target-domain data. The yellow curve shows the performance of the pretrained network when the early layers are frozen and only the later layers are updated during retraining. The shaded bands indicate the statistical uncertainties.}
    \label{fig:JetTaggerResults}
\end{figure}

When adapting to ATLAS Open Data full simulation, the larger domain shift again changes the picture. The results summarized in Figure~\ref{fig:JetTaggerResults} show that freezing the GNN layers consistently degrades performance relative to the baseline, confirming that the pretrained representations are no longer sufficiently flexible. In contrast, allowing all layers to be retrained yields systematic improvements over independent training, with the pretrained model achieving comparable or better performance using roughly half the amount of training data. The model with frozen layers fails to adapt to the domain shift and is therefore unable to capture the relevant differences between the source and target domains.

\subsection{Missing Transverse Energy Reconstruction}

The missing transverse energy reconstruction task constitutes the most challenging scenario studied in this work. It is formulated as a regression problem using a transformer-based architecture and low-level track information, and its performance cannot be fully characterized by a single scalar metric. 
First, the resolution of $E_{\mathrm{T}}^{\mathrm{miss}}$ is a central quantity of interest and is typically defined as the width of the distribution of the difference between the reconstructed and true $E_{\mathrm{T}}^{\mathrm{miss}}$. This resolution is not constant but depends strongly on the true $E_{\mathrm{T}}^{\mathrm{miss}}$ value and the event topology, making it necessary to evaluate performance in different kinematic regimes rather than through a single averaged number. Second, the reconstruction bias, defined as the mean of the same difference distribution, must be controlled independently of the resolution. A model can trivially achieve a small resolution by systematically underestimating or overestimating $E_{\mathrm{T}}^{\mathrm{miss}}$, which would be unacceptable for physics analyses despite a seemingly good scalar metric.
Furthermore, $E_{\mathrm{T}}^{\mathrm{miss}}$ enters analyses both as a continuous observable and through threshold-based selections. This implies that the tails of the reconstructed distributions and the stability of the prediction across the full phase space are as important as the central performance. As a result, a meaningful evaluation of an $E_{\mathrm{T}}^{\mathrm{miss}}$ regression model requires a combination of metrics, including loss values, resolution and bias as functions of true $E_{\mathrm{T}}^{\mathrm{miss}}$, and comparisons of full distribution shapes, rather than reliance on a single scalar performance measure. In this study, a fixed test set is defined for both the CMS Delphes samples and the ATLAS Open Data samples and is kept identical for all training instances to ensure a consistent and unbiased comparison. Model performance is first evaluated using the value of the loss function on this test set as a proxy for overall performance. In a second step, the physical quality of the prediction is assessed by examining the resolution distributions of the reconstructed missing transverse energy, providing a more detailed and physically meaningful evaluation of the model behavior.

\begin{figure}[htbp]
    \centering
    \includegraphics[width=0.49\textwidth]{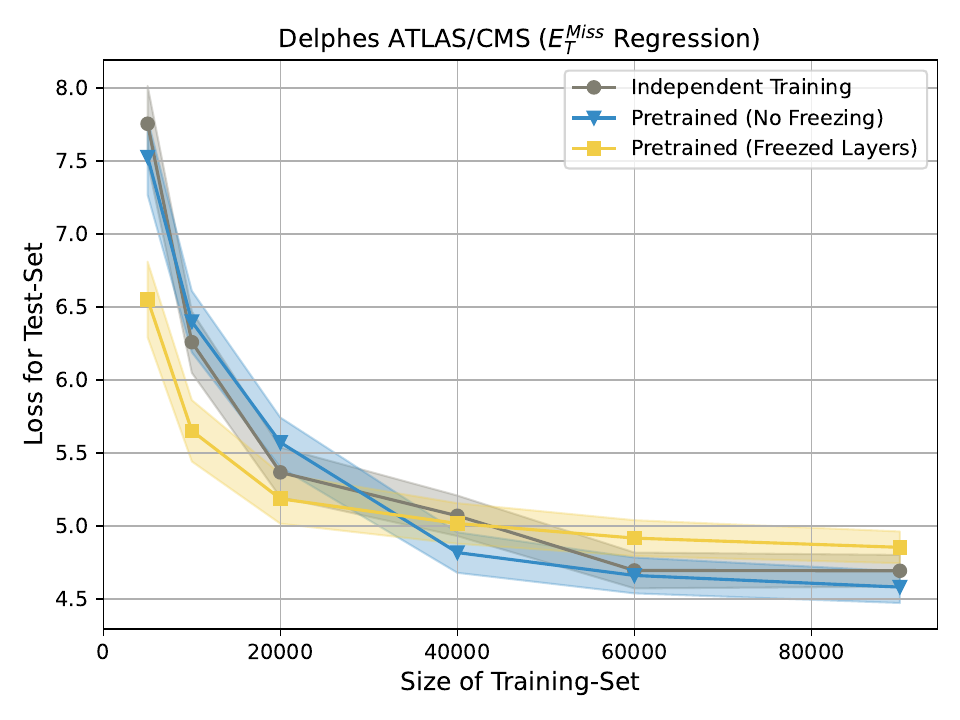}
    \includegraphics[width=0.49\textwidth]{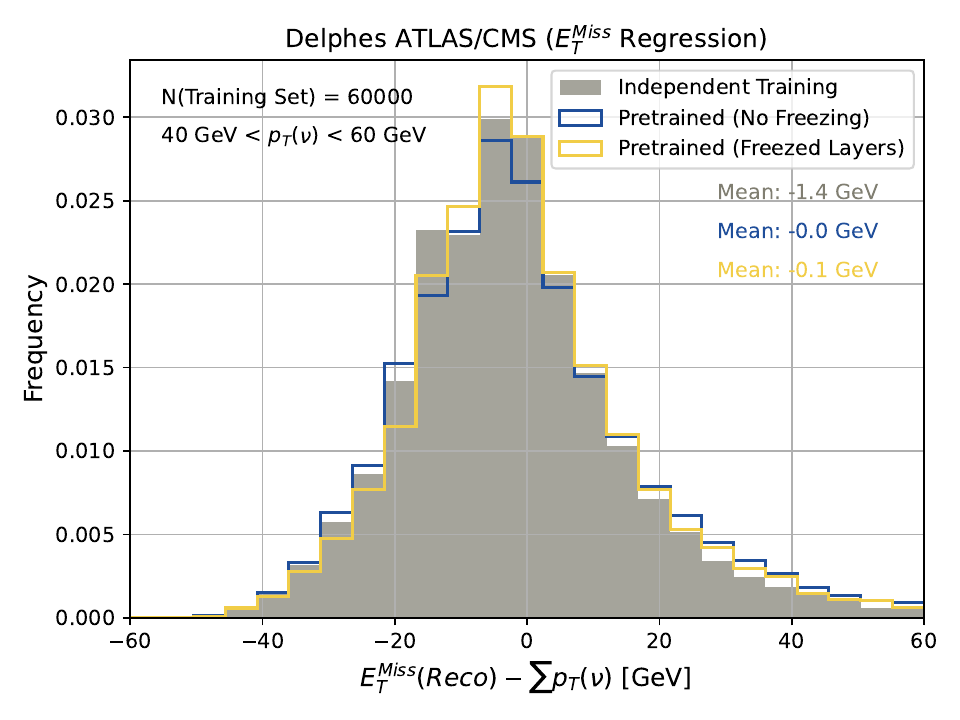}
    \caption{Left: Network performance quantified by the loss function on a test-set as a function of the training set size using the Delphes CMS training samples. Right: The final network performance on the regression task for true $E_T^{Miss}$ values between 40 and 60 GeV. The gray curves shows the performance of a network trained from scratch with randomly initialized weights. The blue curves corresponds to a network pretrained on the ATLAS Delphes samples, where all network parameters are retrained on the target-domain data. The yellow curves shows the performance of the pretrained network when the early layers are frozen and only the later layers are updated during retraining.}
    \label{fig:ETMissResults1}
\end{figure}

For CMS Delphes samples, the evolution of the test-set loss with training set size is shown in Figure~\ref{fig:ETMissResults1}. The pretrained model with full retraining achieves the lowest loss values across the full range of statistics. However, in contrast to the classification tasks, a significantly larger fraction of the target-domain data is required to match the performance of the independently trained baseline. Approximately $70\%$ of the full training sample is needed to reach a comparable loss. The benefit of pretraining becomes evident at larger statistics: when trained on the full CMS Delphes dataset, the pretrained model outperforms the baseline, indicating improved asymptotic performance. However, Freezing the transformer layers leads to inferior performance in all regimes, suggesting that the attention mechanisms require substantial adaptation even for relatively moderate domain shifts.

To complement the loss-based evaluation, the reconstructed $E_{\mathrm{T}}^{\mathrm{miss}}$ resolution is examined. The distributions of the difference between true and reconstructed missing transverse energy for events with $40 < E_{\mathrm{T}}^{\mathrm{miss,true}} < 60$~GeV are also shown in Figure \ref{fig:ETMissResults1}. By construction, all models exhibit only a small bias. 

When transferring to ATLAS Open Data full simulation, the impact of pretraining becomes more pronounced. The loss evolution shown in Figure~7 demonstrates that both pretrained models outperform the independently trained baseline across all training set sizes. In this case, only about half of the available training data is required to achieve a similar performance, despite the large differences in detector simulation and physics modeling. The corresponding resolution distributions, shown in Figure~8, confirm that the improved loss translates into equal or better $E_{\mathrm{T}}^{\mathrm{miss}}$ resolution at fixed bias.

At first glance, it appears counterintuitive that pretraining yields only a modest improvement when transferring between ATLAS and CMS fast simulation, while providing a significantly larger benefit when adapting to fully simulated ATLAS Open Data, despite the latter representing a much larger domain shift. This behavior suggests that the advantage of pretraining is not primarily due to an improved initial performance in the target domain, but rather to a more favorable optimization starting point. While models trained from scratch can quickly adapt to the relatively small detector-level differences between fast-simulation setups, training on fully simulated data is substantially more challenging due to differences in physics modeling and detector response. In this regime, pretraining might provide an effective inductive bias that constrains the learning process to physically meaningful representations, thereby reducing the amount of data required to reach optimal performance, even though the pretrained model itself is not directly accurate in the target domain. Clearly, this aspect warrants a more detailed investigation and will be addressed in future work.

\begin{figure}[htbp]
    \centering
    \includegraphics[width=0.49\textwidth]{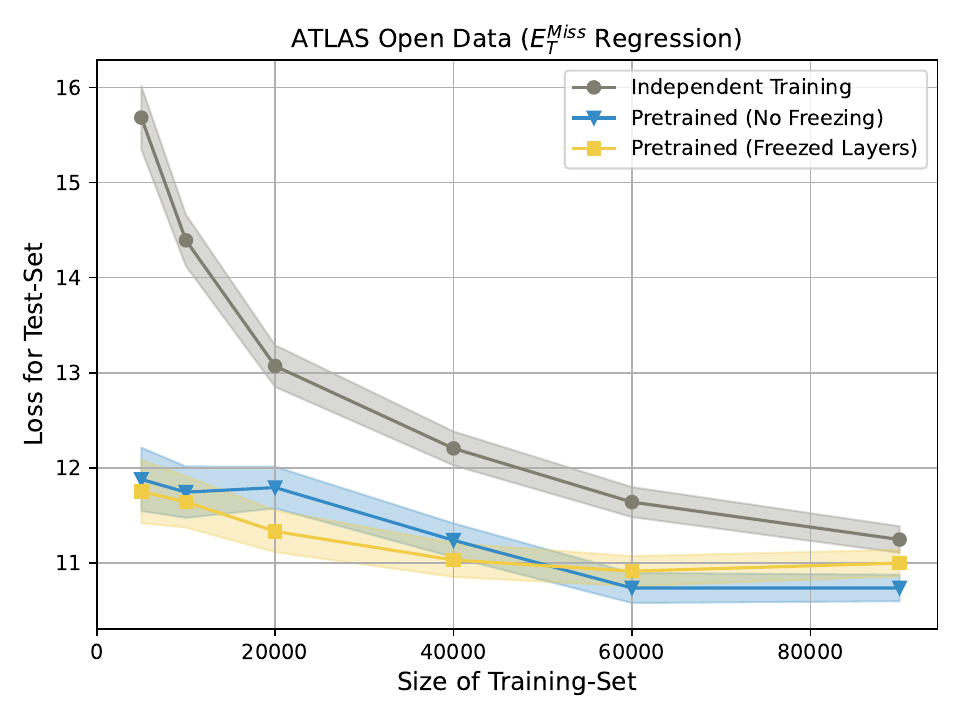}
    \includegraphics[width=0.49\textwidth]{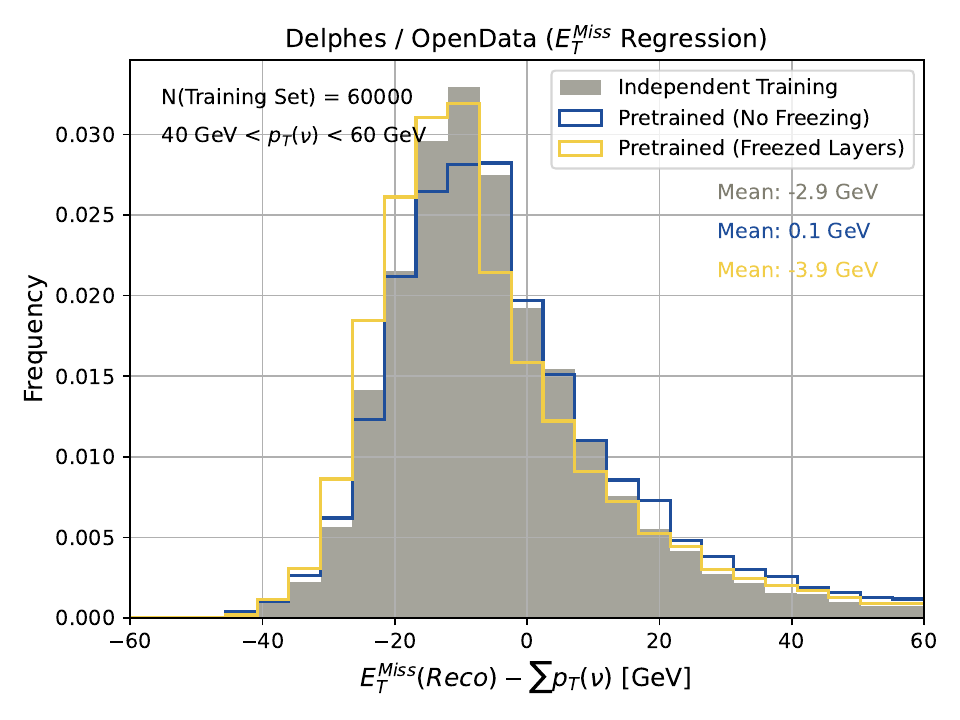}
    \caption{Left: Network performance quantified by the loss function on a test-set as a function of the training set size using the ATLAS Open Data training samples. Right: The final network performance on the regression task for true $E_T^{Miss}$ values between 40 and 60 GeV. The gray curves shows the performance of a network trained from scratch with randomly initialized weights. The blue curves corresponds to a network pretrained on the ATLAS Delphes samples, where all network parameters are retrained on the target-domain data. The yellow curves shows the performance of the pretrained network when the early layers are frozen and only the later layers are updated during retraining.}    \label{fig:ETMissResults2}
\end{figure}

\subsection{Global Trends}

Across all three tasks, several robust conclusions emerge. Pretraining on fast-simulated samples consistently reduces the amount of target-domain training data required to reach a given performance level. This effect is strongest when all network parameters are allowed to adapt to the new domain. Freezing early layers can be advantageous in low-statistics regimes and for moderate domain shifts, but generally limits performance when transferring to fully simulated data. Importantly, these trends hold across very different architectures and input representations, underscoring the broad applicability of transfer learning in contemporary high-energy physics analyses.

\section{Interpretation}

The results presented in Section~4 provide quantitative evidence that transfer learning is an effective strategy across a wide range of machine-learning applications in high-energy physics. While the individual tasks differ substantially in their inputs, architectures, and performance metrics, several unifying patterns emerge that allow a broader interpretation of the findings and their implications for current and future analyses.

\subsection{Impact of Domain Shift and Model Flexibility}

A key observation across all three tasks is the strong dependence of transfer-learning performance on the magnitude of the domain shift between source and target datasets. Transfers between fast-simulated samples with different detector configurations, such as ATLAS Delphes to CMS Delphes, generally benefit from both full and partial retraining strategies. In these cases, freezing early layers can act as a useful inductive bias, particularly in low-statistics regimes, by preserving representations that capture generic kinematic or substructure features.

In contrast, transfers from fast simulation to fully simulated ATLAS Open Data represent a significantly more challenging scenario. Here, differences arise not only from detector response and pile-up modeling but also from the underlying physics description, including matrix-element calculations, parton shower models, and tuning choices. In this regime, freezing early layers consistently degrades performance, while allowing all network parameters to adapt yields systematic improvements over independent training. This demonstrates that pretrained models must retain sufficient flexibility to reconfigure their internal representations when faced with large, qualitative changes in the data distribution.

\subsection{Task Complexity and Transfer Efficiency}

The effectiveness of transfer learning also depends on the intrinsic complexity of the task. For classification problems based on high-level features or structured low-level inputs, such as the signal-background classifier and the quark-gluon jet tagger, pretrained models reach baseline performance with roughly half of the target-domain training data. This suggests that these tasks share common underlying representations that are learned efficiently in the source domain and can be reused with minimal adaptation.

The missing transverse energy reconstruction task, formulated as a regression problem using a transformer-based architecture, exhibits a qualitatively different behavior compared to the classification tasks. While pretraining provides a clear advantage when transferring across the larger domain shift from fast simulation to fully simulated samples, no comparable benefit is observed when transferring only from ATLAS Delphes to CMS Delphes. This behavior likely reflects the increased sensitivity of regression tasks to subtle, domain-specific effects that can be learned efficiently from scratch when the domain shift is small. Nevertheless, even in this more challenging setting, pretraining leads to improved asymptotic performance and significantly reduced requirements on the size of the data-sets when adapting to fully simulated samples, underscoring its value for complex reconstruction tasks.

\subsection{Implications for Fast-Simulation-Based Studies}

A particularly important consequence of these results concerns the interpretation of physics studies that rely primarily, or even exclusively, on fast simulation. A large fraction of phenomenological investigations, feasibility studies, and exploratory analyses, especially those conducted in the context of master’s and bachelor’s theses, are based on fast-simulated samples due to their low computational cost and ease of production. Such studies are often viewed with caution, as it is unclear to what extent conclusions drawn from fast simulation will hold once more realistic detector effects and physics modeling are taken into account.

The present results provide quantitative evidence that this concern can be mitigated when machine-learning models are used in a transfer-learning framework. The fact that pretrained models trained on fast simulation consistently adapt well to fully simulated data, often requiring only limited additional statistics, implies that representations learned in fast simulation capture essential and transferable physical information. Consequently, results obtained in fast-simulation-based studies are more likely to remain valid when confronted with higher-fidelity simulations, provided that appropriate retraining or fine-tuning is performed.

\subsection{Relation to Foundation Models}

At first glance, the ideas explored in this work may appear closely related to the concept of foundation models, which aim to learn universal representations from extremely large and diverse datasets. Indeed, transfer learning is a core principle underlying such models, and recent efforts in high-energy physics have begun to explore large-scale, general-purpose architectures trained on heterogeneous particle-level data.

However, this study addresses a complementary and practically distinct question. Rather than focusing on a single, large, universal model, we investigate whether transfer learning is effective across a broad range of task-specific architectures that are already widely used in HEP analyses. Dense neural networks for tabular data, graph neural networks for jet tagging, and transformer-based models for event-level regression all represent established tools with different inductive biases and computational requirements.

The relevance of this study in the era of foundation models lies in several aspects. First, foundation models are expensive to train and are not yet universally accessible to all analyses or experiments. In contrast, the transfer-learning strategies explored here can be applied immediately using existing models and modest computational resources. Second, many analyses will continue to rely on specialized architectures tailored to specific tasks, where training a full foundation model would be neither practical nor necessary. This work demonstrates that the benefits of transfer learning are not confined to large, monolithic models, but also apply to lightweight, task-specific networks.

\subsection{Cross-Experimental Training as a Strategy to Mitigate Data Limitations}

It was recently argued \cite{Vigl:2026ppx} that achieving ultimate jet flavour tagging performance may require training on extremely large datasets, potentially exceeding the statistics expected even after the High-Luminosity LHC. This observation raises an important practical question: how can the effective training statistics be increased without relying exclusively on ever-growing single-experiment datasets?

One natural possibility is to combine fully simulated datasets from multiple experiments. Instead of training exclusively on ATLAS or CMS fully simulated samples, one can exploit the fact that both experiments simulate the same underlying physics processes with different but similar detector realizations. From a machine-learning perspective, this increases the diversity of detector responses and effectively enlarges the accessible training domain.

To explore this idea in a proof-of-principle study, we trained the previously described GNN for quark-gluon jet tagging as well as the transformer-based network for $E_T^{\text{miss}}$ regression. For each model, we compared two training strategies: First, training exclusively on ATLAS Open Data, second, training on a combined dataset of equal total size, composed of 50\% ATLAS and 50\% CMS Open Data (both fully simulated). Importantly, the total number of training events was kept fixed in both cases. For example, a training size of 10k events in the combined configuration corresponds to 5k ATLAS and 5k CMS events.

The dataset structure follows the setup described previously, but is restricted to $W \rightarrow \mu\nu$ samples with comparatively low jet transverse momenta, given by the available data-sets on the Open-Data Portal. Furthermore, the transverse impact parameter $d_0$ was partially masked and no $z_0$ information was available during training, reducing the amount of track-level discrimination information and making the setup more challenging, also yielding to differences to the results discussed before. 

\begin{figure}[htbp]
    \centering
    \includegraphics[width=0.49\textwidth]{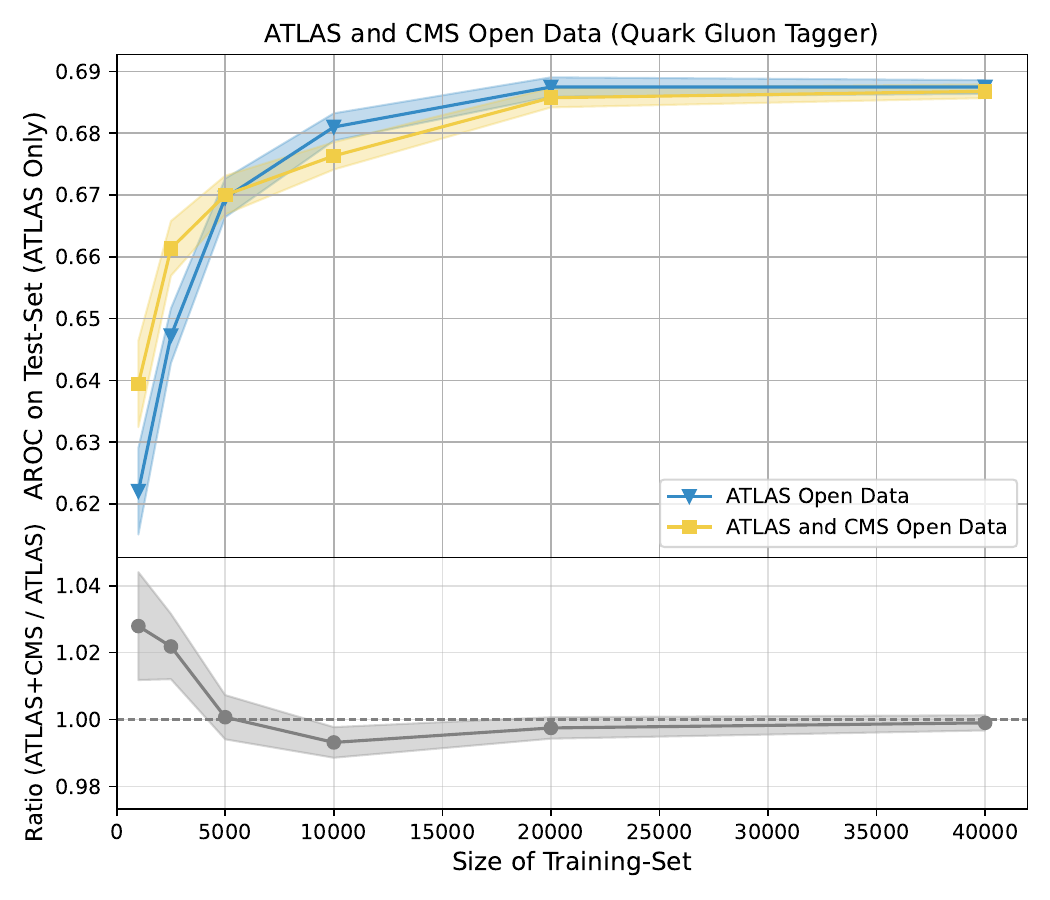}
    \includegraphics[width=0.49\textwidth]{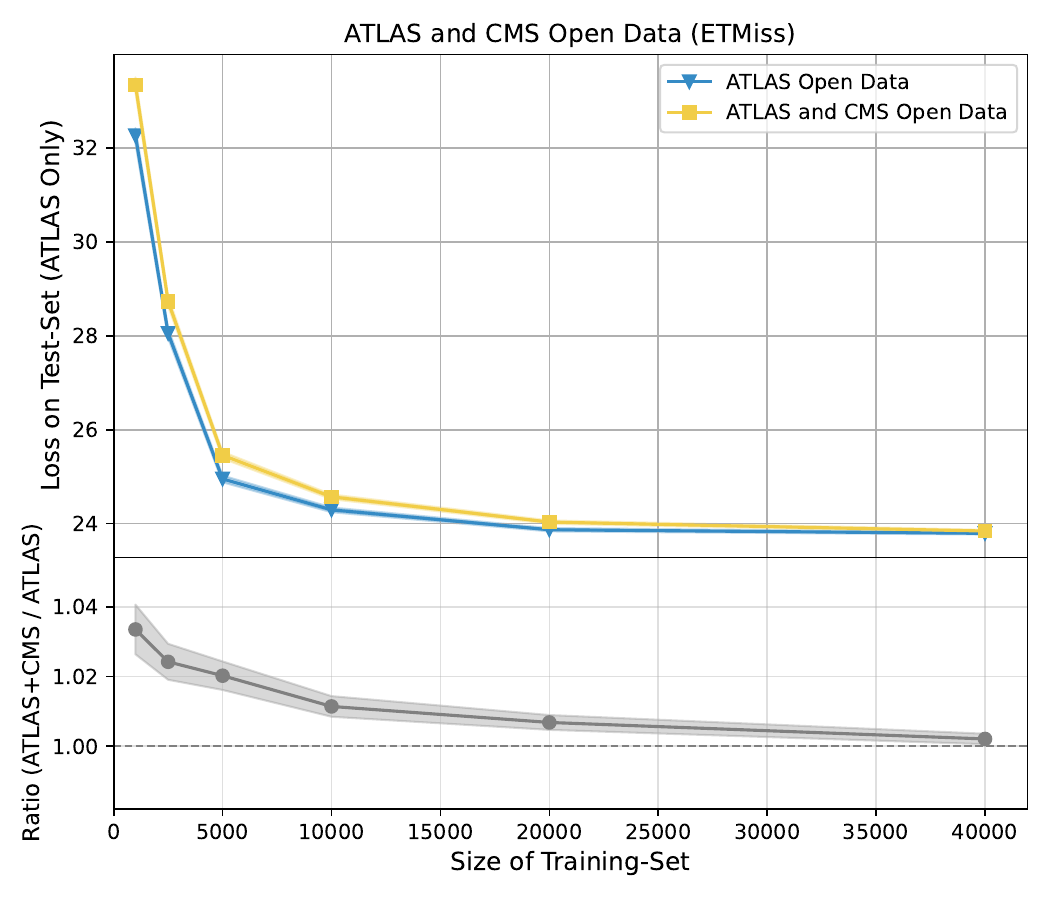}
    \caption{Dependence of the AROC performance of the quark–gluon jet tagger (left) and the test-set loss for the $E_T^{\text{miss}}$ regression (right) as a function of the training set size.
The blue curve shows results obtained using only the ATLAS Open Data sample. The yellow curve shows results obtained from a combined ATLAS+CMS Open Data sample with the same total number of training events. For example, a training size of 10k in the combined setup corresponds to 5k events from ATLAS and 5k events from CMS.}    \label{fig:OpenData2}
\end{figure}

\begin{table}[htbp]
\centering
\begin{tabular}{c | cc | cc}
\hline
 & \multicolumn{2}{c}{\textbf{Quark--Gluon Tagger (AROC)}} & \multicolumn{2}{|c}{\textbf{$E_T^{\text{miss}}$ Regression (Loss on Test Dataset)}} \\
\hline
Training Size & ATLAS & ATLAS+CMS & ATLAS & ATLAS+CMS \\
\hline
1000  & $0.62 \pm 0.01$ & $0.64 \pm 0.01$ & $32.3 \pm 0.2$ & $33.3 \pm 0.2$ \\
2500  & $0.65 \pm 0.01$ & $0.66 \pm 0.01$ & $28.1 \pm 0.2$ & $28.7 \pm 0.2$ \\
5000  & $0.67 \pm 0.01$ & $0.67 \pm 0.01$ & $25.0 \pm 0.2$ & $25.5 \pm 0.2$ \\
10000 & $0.68 \pm 0.01$ & $0.68 \pm 0.01$ & $24.3 \pm 0.2$ & $24.6 \pm 0.2$ \\
20000 & $0.69 \pm 0.01$ & $0.69 \pm 0.01$ & $23.9 \pm 0.2$ & $24.0 \pm 0.2$ \\
40000 & $0.69 \pm 0.01$ & $0.69 \pm 0.01$ & $23.8 \pm 0.2$ & $23.8 \pm 0.2$ \\
\hline
\end{tabular}
\caption{Performance as a function of training dataset size for two representative tasks. The left columns show the area under the ROC curve (AROC) for the quark--gluon jet tagger, while the right columns show the test-set loss for the $E_T^{\text{miss}}$ regression. Results are given for models trained on ATLAS Open Data only and for models trained on a combined ATLAS+CMS Open Data sample of equal total size.}
\label{tab:combined_training_size}
\end{table}

The results are detailed in Table \ref{tab:combined_training_size} and summarized in Fig.~\ref{fig:OpenData2}. The left panel shows the AROC performance of the quark--gluon tagger evaluated on the ATLAS test set as a function of training size. The right panel shows the corresponding test loss for the $E_T^{\text{miss}}$ regression task, again purely on an ATLAS based test set. In both cases, the upper subpanels display the absolute performance, while the lower subpanels show the ratio between the mixed (ATLAS+CMS) and ATLAS-only training. It should be noted that the mixed datasets are expected to show slightly worse performance than the ATLAS-only training when comparing the same number of events, since only half of the events originate from the ATLAS simulation used for evaluation. The critical question, however, is whether a mixed dataset of a given size can outperform an ATLAS-only training with a smaller number of events. Such a comparison indicates whether the inclusion of additional CMS data provides a net benefit by effectively increasing the usable training statistics.

For the quark--gluon tagger, the mixed training strategy yields comparable or slightly improved performance at small training sizes, while converging to similar values at larger statistics. For the $E_T^{\text{miss}}$ regression, a similar trend is observed: the combined dataset leads to competitive performance across all training sizes, with the ratio approaching unity as the dataset grows. The ratio panels indicate that cross-experimental training does not degrade detector-specific performance, despite the inclusion of events from a different detector simulation. While the ATLAS+CMS data-set yields a partially a slightly worse performance it still adds in all cases significant information.

Although this study is limited in scope, both in model size and in total available statistics, and therefore does not yet allow for full quantitative  conclusions, it demonstrates a promising direction. Combining fully simulated datasets across experiments may provide indeed an effective strategy to increase the diversity and utility of training data, potentially alleviating the extreme data requirements suggested for next-generation flavour tagging performance. A dedicated large-scale study using modern high-capacity architectures and substantially larger datasets will be required to assess the full impact of this approach.

\subsection{Implications for Model Sharing and Reproducibility}

An important methodological implication of this work is the way machine-learning results should be shared within the HEP community. If pretrained models trained on fast simulation provide a strong and reusable starting point across experiments, detector configurations, and physics models, then publishing only the model architecture or training code is no longer sufficient. The trained weights themselves become a central scientific artifact.

The results presented here strongly motivate the systematic publication of pretrained models alongside analysis code. Making trained models publicly available would enable other studies to directly build upon existing work, reduce redundant training efforts, and facilitate meaningful comparisons across analyses. This applies not only to large foundation models, but equally to task-specific networks such as classifiers, taggers, and regressors that are routinely used in physics analyses.

\section{Summary and Outlook}

In this paper, we have presented a systematic study of transfer learning for machine-learning applications in high-energy physics, spanning three representative tasks: signal-background classification, quark-gluon jet tagging, and missing transverse energy reconstruction. These tasks were deliberately chosen to cover a broad range of use cases, input representations, and neural network architectures commonly employed in contemporary LHC analyses.

Using fast-simulated samples as a source domain and both alternative fast simulations and fully simulated ATLAS Open Data as target domains, we demonstrated that pretrained models typically outperform independently trained models when adapted to new simulation environments. Across all tasks, pretraining reduces the amount of target-domain training data required to reach a given performance level, often by a factor of two. The benefit is most pronounced when all model parameters are allowed to adapt to the new domain, while freezing early layers is only advantageous for moderate domain shifts and low-statistics regimes.

A particularly important outcome of this study is the demonstration that models trained on fast-simulated data learn representations that remain useful even when confronted with substantially different detector simulations and physics modeling. This provides quantitative support for the validity of many physics studies that rely on fast simulation, especially in exploratory analyses and student projects, provided that transfer learning is used when moving toward more realistic simulations.

Looking ahead, these results motivate a broader shift in how machine-learning models are developed and shared within the high-energy physics community. Rather than treating trained networks as disposable by-products of individual analyses, pretrained models should be regarded as reusable scientific assets. The systematic publication of trained weights, alongside code, datasets, and documentation, would enable more efficient use of simulation resources, facilitate cross-experiment collaboration, and accelerate the development of new analyses.

While large foundation models represent an important and complementary direction, this work demonstrates that transfer learning is already highly effective for task-specific architectures that are widely used today. Extending these studies to additional physics processes, detector configurations, and real collision data will be an important next step. Ultimately, integrating transfer learning into standard analysis workflows has the potential to significantly improve the sustainability, reproducibility, and scientific reach of machine-learning-based analyses in high-energy physics.

\section*{Acknowledgements}
We acknowledge the work of the ATLAS Collaboration to record or simulate, reconstruct, and distribute the Open Data used in this paper, and to develop and support the software with which it was analysed.

\bibliographystyle{unsrt}
\bibliography{./Bibliography}

\end{document}